\pdfoutput=1
\documentclass[10pt, conference]{IEEEtran} 
\usepackage[left=0.75in,right=0.75in,top=0.75in,bottom=0.75in]{geometry}
\IEEEoverridecommandlockouts

\usepackage[utf8]{inputenc} 
\usepackage[T1]{fontenc}    
\usepackage{hyperref}       
\usepackage{url}            
\usepackage{booktabs}       
\usepackage{amsfonts}       
\usepackage{nicefrac}       
\usepackage{microtype}      
\usepackage{xcolor}         
\usepackage{amsmath,amssymb,multicol,latexsym}
\usepackage{pgfplots,pgfplotstable}
\usetikzlibrary{pgfplots.groupplots}
\usepackage{subcaption}
\pgfplotsset{compat=1.16}
\usepackage{tikz}
\usetikzlibrary{external}
\usepackage{hyperref}
\usepackage{cleveref}
\usepackage{todonotes}
\usepackage{url}
\usepackage{graphicx}
\usepackage{svg}
\usepackage{cite}
\usepackage{flushend} 

\graphicspath{{figures/}}

\makeatletter
\def\ps@IEEEtitlepagestyle{%
  \def\@oddfoot{\mycopyrightnotice}%
  \def\@oddhead{\hbox{}\@IEEEheaderstyle\leftmark\hfil\thepage}\relax
  \def\@evenhead{\@IEEEheaderstyle\thepage\hfil\leftmark\hbox{}}\relax
  \def\@evenfoot{}%
}
\def\mycopyrightnotice{%
  \begin{minipage}{\textwidth}
  \centering \scriptsize
  Copyright~\copyright~2024 IEEE. Personal use of this material is permitted. Permission from IEEE must be obtained for all other uses, in any current or future media, including reprinting/republishing this material for advertising or promotional purposes, creating new collective works, for resale or redistribution to servers or lists, or reuse of any copyrighted component of this work in other works.
  \end{minipage}
}
\makeatother

\title{\vspace{0.25in}Scene-Extrapolation: Generating Interactive Traffic Scenarios}

\author{
Maximilian~Zipfl$^{1,2}$,
Barbara~Schütt$^{1,2}$, 
J.~Marius~Zöllner$^{1,2}$
\thanks{$^{1}$FZI Research Center for Information Technology, Karlsruhe, Germany
{\tt\small \{zipfl, schuett, zoellner\}@fzi.de}}%
\thanks{$^{2}$KIT - Karlsruhe Institute of Technology, Karlsruhe, Germany}\\%
\thanks{The research leading to these results is funded by the German Federal Ministry for Economic Affairs and Climate Action within the project “Verifikations- und Validierungsmethoden automatisierter Fahrzeuge im urbanen Umfeld”. The project is a part of the PEGASUS family. The authors would like to thank the consortium for the successful cooperation.}
}

\date{January 2024}

\begin{document}

\maketitle

\begin{abstract}
Verifying highly automated driving functions can be challenging, requiring identifying relevant test scenarios. Scenario-based testing will likely play a significant role in verifying these systems, predominantly occurring within simulation.
In our approach, we use traffic scenes as a starting point (seed-scene) to address the individuality of various highly automated driving functions and to avoid the problems associated with a predefined test traffic scenario. Different highly autonomous driving functions, or their distinct iterations, may display different behaviors under the same operating conditions.
To make a generalizable statement about a seed-scene, we simulate possible outcomes based on various behavior profiles.
We utilize our lightweight simulation environment and populate it with rule-based and machine learning behavior models for individual actors in the scenario.
We analyze resulting scenarios using a variety of criticality metrics. The density distributions of the resulting criticality values enable us to make a profound statement about the significance of a particular scene, considering various eventualities.
\end{abstract}


\section{Introduction}

The large-scale introduction of highly automated vehicles (HAV) on public roads is the next big step in the transport sector. One significant challenge is verifying and validating the safety of these vehicles, with behavior planning being one of the most crucial and complex components.
The European Commission has passed a law \cite{ryglewski_verordnung_2022, european_commission_commission_2022} to regulate the safety of HAVs. This law requires that highly automated driving functions undergo scenario-based testing for verification purposes. Not all essential test scenarios have been fully defined yet, which raises the question of which scenarios should be prioritized and included in the testing process.
Scenario-based testing is a concept in which testers challenge the System Under Test (SUT) by executing a range of discrete runs, known as scenarios, and evaluate the quality of each run based on given metrics. The main challenge in scenario-based testing is selecting the appropriate test scenarios.
The greater the number of scenarios, the better the transferability of the tested scenarios to real-life behavior.
However, complete coverage is impossible due to the potentially infinite number of scenarios. Instead, testing must be concluded at a certain point in time, to make a statement regarding the safety of the HAV.

Numerous approaches to scenario-based testing have been investigated in recent years. These approaches can generally be categorized as knowledge-driven and data-driven scenario generation \cite{riedmaier_survey_2020}.
The knowledge-driven approach involves extracting information from various sources, such as existing scenarios, functional descriptions, expert opinions, traffic guidelines, and standards. This information is then structured and composed into new scenarios using combinatorial techniques \cite{riedmaier_survey_2020, wotawa_ontology-based_2020, klueck_using_2018}.
The data-driven, alternative method relies on known scenarios from accident databases or real road traffic. Unlike knowledge-based methods, the data-driven approach provides increased realism by avoiding limitations and biases associated with expert models \cite{ries_trajectory-based_2021}. 
\begin{figure}[t]
    \centering
    \def\svgwidth{0.96\columnwidth}
\begingroup%
  \makeatletter%
  \providecommand\color[2][]{%
    \errmessage{(Inkscape) Color is used for the text in Inkscape, but the package 'color.sty' is not loaded}%
    \renewcommand\color[2][]{}%
  }%
  \providecommand\transparent[1]{%
    \errmessage{(Inkscape) Transparency is used (non-zero) for the text in Inkscape, but the package 'transparent.sty' is not loaded}%
    \renewcommand\transparent[1]{}%
  }%
  \providecommand\rotatebox[2]{#2}%
  \newcommand*\fsize{\dimexpr\f@size pt\relax}%
  \newcommand*\lineheight[1]{\fontsize{\fsize}{#1\fsize}\selectfont}%
  \ifx\svgwidth\undefined%
    \setlength{\unitlength}{399.79918472bp}%
    \ifx\svgscale\undefined%
      \relax%
    \else%
      \setlength{\unitlength}{\unitlength * \real{\svgscale}}%
    \fi%
  \else%
    \setlength{\unitlength}{\svgwidth}%
  \fi%
  \global\let\svgwidth\undefined%
  \global\let\svgscale\undefined%
  \makeatother%
  \begin{picture}(1,0.54746065)%
    \lineheight{1}%
    \setlength\tabcolsep{0pt}%
    \put(0,0){\includegraphics[width=\unitlength,page=1]{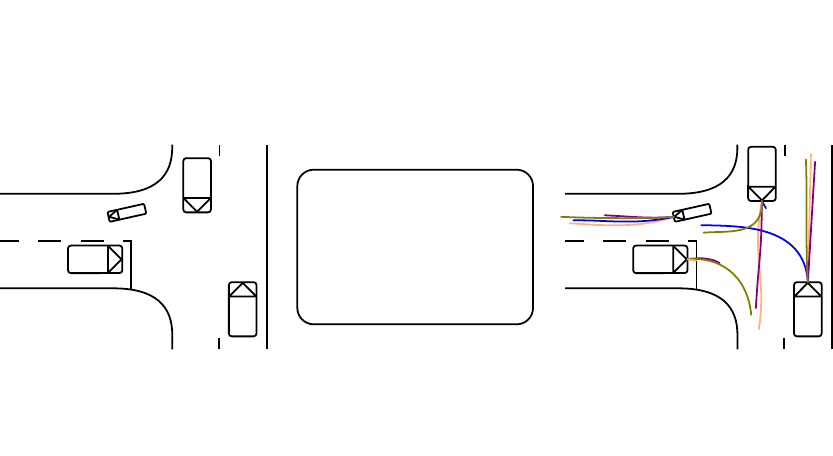}}%
    \put(0.49912066,0.25409028){\color[rgb]{0,0,0}\makebox(0,0)[t]{\smash{\begin{tabular}[t]{c}Simulation\\Framework\end{tabular}}}}%
    \put(0.44037156,0.50105672){\color[rgb]{0,0,0}\makebox(0,0)[t]{\smash{\begin{tabular}[t]{c}Model Collection\end{tabular}}}}%
    \put(0,0){\includegraphics[width=\unitlength,page=2]{resimulation_framework_scheme.pdf}}%
    \put(0.48214886,0.39491196){\color[rgb]{0,0,0}\makebox(0,0)[t]{\smash{\begin{tabular}[t]{c}\small{Model}\end{tabular}}}}%
    \put(0,0){\includegraphics[width=\unitlength,page=3]{resimulation_framework_scheme.pdf}}%
    \put(0.49952187,0.03648142){\color[rgb]{0,0,0}\makebox(0,0)[t]{\smash{\begin{tabular}[t]{c}Criticality Assessment\end{tabular}}}}%
    \put(0,0){\includegraphics[width=\unitlength,page=4]{resimulation_framework_scheme.pdf}}%
    \put(0.16836006,0.44008862){\color[rgb]{0,0,0}\makebox(0,0)[t]{\smash{\begin{tabular}[t]{c}Seed-Scene\end{tabular}}}}%
  \end{picture}%
\endgroup%

    \caption{Scene extrapolation: possible futures of a seed-scene are simulated by applying different permutations of behavior models. The resulting child-scenarios are assessed by criticality metrics.}
    \label{fig:approach_topright}
    \vspace{-.05cm}
\end{figure}
Simulation is considered the most promising tool for realizing scenario-based testing.
To ensure the validity and relevance of a simulation, it is crucial that traffic scenario entities show reactive behavior. This involves accurately reflecting how different traffic participants dynamically respond to their environment.
In complex traffic scenarios, a hybrid simulation that combines recorded trajectories of traffic participants with intelligent actors is usually unsuccessful \cite{bernhard_bark_2020}.
Consequently, it is essential to employ a dynamically responsive testing environment to assess and compare the driving capabilities of various HAVs. Consider a scenario involving a merging vehicle under event and time-based control. Depending on the vehicle's speed, a specific driving function may be deemed critical in one instance, while another driving function, having already passed the on-ramp, may be deemed irrelevant. 
In addition, searching for relevant parameters for testing can be computationally intensive, particularly when many traffic participants are involved in a scenario. Additionally, as seen in the example, a supposedly critical scenario may turn out to be insignificant if the SUT operates outside a certain parameter range.
This highlights the need for a dynamic, interactive testing scenario to evaluate individual HAV performance accurately.

In this paper, we propose an approach to generate testing scenarios which differs from conventional methods, maintaining independence from the driving function being tested. Instead of relying on defined scenarios, we use seed-scenes. The seed-scenes only specify the starting point of each scenario.
The actual course of the scenario is determined by each individual driving function that is to be tested, along with the behavior profiles of the other traffic participants selected by the test engineer.
We extrapolate the scenes into different futures in a closed-loop simulation based on the behavior models of the actors involved, to provide a general understanding of the seed-scene. A comprehensive evaluation of all simulated futures should then provide a measure of a scene's importance for testing HAVs.

The contributions of this work are as follows:
\begin{itemize}
    \item  Development and implementation of a simulator for the resimulation of traffic scenarios based on a seed-scene and various behavior models
    \item  Concept for the evaluation of the mentioned seed-scenes with regard to the testing of HAV
\end{itemize}

This paper is further structured as follows. \Cref{sec:related_work} presents related work and discusses its commonality with our approach. Here, we focus on the aspects of the simulator and the different behavior models. In \Cref{sec:framework}, we present our methodology in detail, introducing the simulator developed, the dataset used, and the behavior models utilized. We then outline and discuss the experiments in \Cref{sec:experiments}. We conclude the work in \Cref{sec:conclusion} and identify possible next steps.

\section{Related Work}
\label{sec:related_work}

\subsection{Simulators}
Simulators have been investigated for some time as a tool for the development and training of highly automated driving functions (HADF). Different simulators are available for different applications, taking into account specific features and environments. These include macroscopic traffic flow simulation (e.g. SUMO \cite{krajzewicz_sumo_2002})  through to physically based real-world simulation. In recent years, a wide variety of simulators have been introduced, particularly in the field of computer vision. Some of the most well-known simulators include GTA V \cite{richter_playing_2017}, Carla \cite{dosovitskiy_carla_2017}, and its extension SUMMIT \cite{cai_summit_2020}. These simulators generate photorealistic environments of traffic scenarios. These are particularly suitable for testing complete, highly automated systems, including sensor perception.

Our research utilizes a sophisticated multi-agent simulator to create a seed-scene based on real-world scenarios. Recent advancements in simulation technologies and benchmarks have significantly influenced our work. Prominent examples include BARK \cite{bernhard_bark_2020}, Nocturne \cite{vinitsky_nocturne_2023}, nuPlan \cite{caesar_nuplan_nodate}, and Waymax \cite{gulino_waymax_2023}, each of which shares similarities with our simulator. These state-of-the-art simulators are adept at supporting multi-agent simulations by utilizing authentic input data, employing a 2D coordinate system to ensure a streamlined and lightweight design.

Several studies have recently been carried out on the generation of traffic scenes or traffic scenarios.
Tan et al. propose an autoregressive approach, SceneGen \cite{tan_scenegen_2021}, to address the limitations of manual and rule-based methods in representing weld complexity. SceneGen iteratively adds various traffic participants to an initial scene for accurate representation.
Feng et al.'s work, TrafficGen \cite{feng_trafficgen_2023}, builds upon SceneGen's principles, employing an autoregressive machine learning model for scene augmentation and new scene generation. It integrates a trajectory prediction module and utilizes a reinforcement simulator to train a behavior predictor for traffic participants controlled by IDM and lane change models.
Similarly, TrafficSim \cite{suo_trafficsim_2021} aims for realistic multi-agent behavior generation using a combined latent scenario representation. The concept behind this work is comparable to the scenario generation proposed in our study. However, unlike our approach, TrafficSim depends on a single learned model to describe the behavior of all traffic participants, which provides inherent knowledge of their behavior.

It is noteworthy that the primary focus of our environment modeling is directed toward the evaluation of the behavioral aspects of both the ego vehicle and other vehicles within the simulated scenarios. In contrast to prioritizing the perception capabilities of HAVs, our approach does not require sensor modeling. The simulators being considered focus on providing behavior-specific features, such as the exact position, size, and path of vehicles moving within the simulated road network. This emphasis on behavioral complexity contributes to a more sophisticated and targeted analysis of vehicle interactions within our simulation framework.

\subsection{Driving Behavior Models}
Machine learning approaches are increasingly popular in the context of automated driving \cite{casas_intentnet_2021, chai_multipath_2019}. This includes reinforcement-based closed loop approaches, which are made possible by the increasing accessibility and popularity of traffic simulators, as well as the increasing variety of traffic datasets \cite{zhan_interaction_2019, caesar_nuscenes_2020, bock_ind_2019, chang_argoverse_2019} that are available to the public.
Reinforcement approaches \cite{isele_navigating_2018, kiran_deep_2022} aim to function effectively in an environment based on their actions. The reward evaluates each action or sequence of actions. Through multiple training iterations, a model becomes capable of handling a specific environment, such as a traffic scene. However, to achieve this, a simulation environment must first be available for training the model. In this work, we aim to incorporate reactive behavior into the simulation environment so we cannot train them in the first place. Therefore, we will also focus on imitation learning. This approach involves cloning the behavior of real drivers, using their trajectories as a reference. Typically, these expert trajectories are derived from real datasets.
In recent years, there has been an increasing use of graph-based approaches \cite{gao_vectornet_2020, zhao_tnt_2020, diehl_graph_2019, zhou_ast-gnn_2021,mo_graph_2021, li_grip_2020, ma_multi-agent_2021, park_leveraging_2023}. It has been found that approaches which have concrete relations between entities, in the form of a graph, can effectively map social interactions between traffic participants and driving behavior. Therefore, this work will also utilize graph-based prediction models \cite{zipfl_relation-based_2022, zipfl_utilizing_2023}.

\section{Simulation Framework}

The simulation framework aims to evaluate scenes for testing highly automated driving. \Cref{fig:approach_topright} presents a schematic of the entire framework. The input is the seed-scene, which can be created from real recordings or synthetically. 
A seed-scene comprises a road map containing information about the geometric and topological properties of the road network. Furthermore, the position, orientation, and speed are provided for each traffic participant. 
Depending on the behavior model applied later, the history of the traffic participants can be used in addition to the current state of the traffic scene. Therefore, a time series can be included at the initial time step of the scene extrapolation.
During the simulation stage, which we refer to in this paper as scene extrapolation, the seed-scene is used to simulate possible futures, which we call child-scenarios. Realistic behavior models are randomly assigned to each actor for every simulation run.

We believe that it is essential for the evaluation of the scenes and the resulting future behavior that individual actors can react to the environment. Therefore, every future scenario must be simulated as a closed-loop.
Criticality metrics are used to evaluate each simulated future, and the calculated results are then summarized to indicate the criticality of the seed-scene. This method provides insight into the relevance of the seed-scene for testing HAVs.

\subsection{Criticality Metrics}
\label{sec:metrics}
In contrast to most research on criticality metrics and scenario-based testing \cite{steimle_toward_2022}, we do not have a designated ego vehicle in our scenes.
However, most metrics are measured between two traffic participants, e.g., between an ego vehicle and a second vehicle nearby.
Also, more general metrics that do not require two traffic participants at least need one point-of-view vehicle to which the related traffic is measured.
For the metric computation, we distinguish between nanoscopic (i.e., metrics that are calculated every time step) and microscopic (i.e., metrics with aggregated values over time or over a scene) \cite{schutt_taxonomy_2022}.
We compute the nanoscopic criticality metric for each vehicle in each scene, i.e., for each vehicle, the criticality is assessed regarding each other vehicle visible in the scene, to get an objective scene criticality evaluation.
These nanoscopic metrics can be aggregated to microscopic scene criticality \cite{schutt_taxonomy_2022} to achieve a better comparison among different scenes.
Depending on what shall be compared, this can be the measured mean value of a criticality metric or its maximum or minimum values within a scene \cite{zipfl_fingerprint_2022}. To increase  information quality of a scenario, we also add the mean value of all maxima of scenes in the scenario for each metric.
We used the following metrics for our assessment:
\begin{itemize}
    \item \textbf{Distance}: Euclidean distance between the centers of two traffic participants.
    \item \textbf{Traffic Quality (TQ)}: The overall assessment of the traffic within the scene where a vehicle is situated \cite{zipfl_fingerprint_2022, schutt_inverse_2023}.
    \item \textbf{Gap Time (GT)}: The predicted distance in time between the two traffic participants crossing an intersection point \cite{allen_brian_l_analysis_1978}.
        \item \textbf{Time-to-collision (TTC)}: Minimal time two traffic participants in a car-following scenario need until a collision. Is case the leading is faster, TTC approaches infinity. Therefore, we apply the inverse TTC \cite{tamke_flexible_2011}.
    \item \textbf{Potential time-to-collision (PTTC)}: TTC with the constraint, that the leading vehicle is decelerating and the following vehicle moves with constant velocity \cite{wakabayashi_traffic_2003}.
    \item \textbf{Worst time-to-collision (WTTC)}: TTC without the limitation to car-following scenarios \cite{wachenfeld_worst-time--collision_2016}.
\end{itemize}

\subsection{Dataset}

The pipeline described in this paper is heavily based on real recorded traffic scenes. Both seed-scenes and machine-learned traffic models presented later in the paper (see \Cref{sec:behavior_models}) are derived from real data.
The INTERACTION dataset \cite{zhan_interaction_2019} is used both for training the models and for analyzing and selecting the seed-scenes.
This dataset focuses on vehicle behavior on roads in three different countries at eleven different locations, covering a variety selection of traffic behaviors and scenes.
Each entity is characterized by its coordinates in Cartesian space, a categorization, and a speed recorded at regular intervals of 10\,Hz. Additionally to trajectories of the objects, associated road information is supplied in the form of high-definition maps. 
The dataset is split into cases, each containing 40 time steps.
This dataset only includes vehicles in road traffic and does not cover other traffic participant types. The work does not consider behavioral models for pedestrians.

In general, however, the simulator's data interface has been designed so that other motion datasets can also be read in. The prerequisite is that at least the attributes mentioned above are included for each traffic participant and that a high-definition map is provided.

\subsection{Simulator}
\label{sec:framework}
\begin{figure*}[htb!]
    \centering
    \def\svgwidth{\textwidth}
    \small{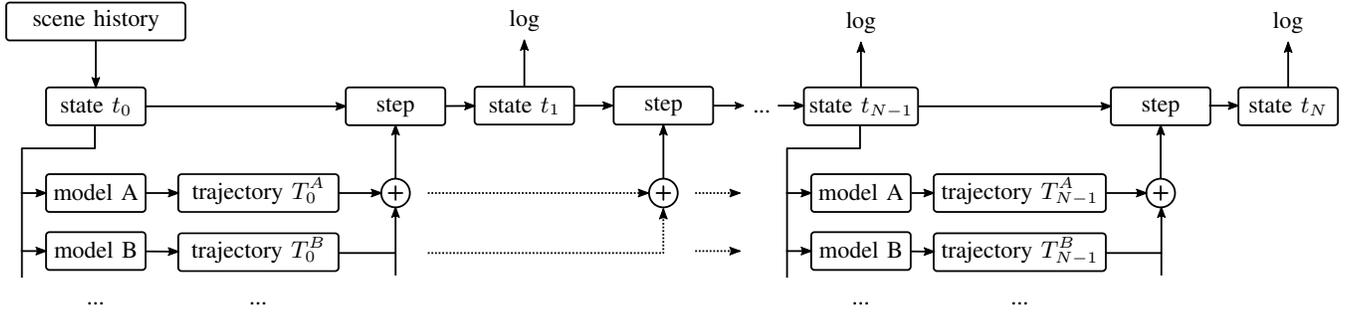}
    
    \caption{Block diagram illustrating the simulation process. Trajectory proposals are computed from a seed-scene, considering various behavior models based on the current situation. Each trajectory is followed for a certain timespan before a recalculation is performed, taking into account the changed scene context.}
    \label{fig:simulatorfig}
\end{figure*}

The simulator functions as an execution unit, passing an input scene to various behavior models and updating calculated trajectories to the next state after each step. Interfaces between modules are standardized, allowing for module exchange as desired. Other simulators, such as those presented in \Cref{sec:related_work}, can also be employed if additional aspects of the environment need to be simulated. Assuming a physics calculation is required, CARLA \cite{dosovitskiy_carla_2017} may be used instead of the simulation framework implemented for this work. This paper focuses solely on behavior and resulting trajectories, which can only be calculated using object lists and road maps depending on the model. Therefore, our simulator can be implemented in a lightweight manner.

\Cref{fig:simulatorfig} illustrates the sequence of simulation steps. Initially, a traffic scene to be extrapolated is selected. 
Furthermore, behavior models are assigned to each traffic participant in the seed-scene. This can be done manually or randomly using a distribution function.
The simulator provides environment and context information for individual models. In line with current research in the field of trajectory prediction, it also includes a history of the last second. This allows for an easy integration of additional imitation learning models.

The simulator provides the same environment for each trajectory calculation step, allowing the models to be computed in parallel. This environment includes the pose, dimensions, speed, and classification (type) of all traffic participants for the current and previous 10 time steps. Additionally, a high-definition road map containing the road geometry and topology is provided.
Using this contextual information, the models calculate a trajectory for the assigned traffic participants.
The models not only compute the next state of their own vehicle, but a planned trajectory for at least the next 30 steps. The simulation frequency is based on the INTERACTION dataset and is 10\,Hz.
To improve simulation performance, trajectory calculation can be skipped for a parametrizable number of steps, allowing actors to follow previously calculated trajectories for a short period of time. The simulation yields best results when calculating trajectories every five to ten steps. These findings align with those presented in \cite{gulino_waymax_2023}.
The simulation duration for a scenario is on average 0.5\,s, which is highly dependent on the choice of behavioral models, with the inference of the machine learning models being the limiting factor.
After each step, the states of all actors are consolidated and recorded in the logger. The data format for storing scenarios follows the specifications of the INTERACTION dataset, ensuring sustainable reusability and the use of other toolkits. 

It is important to note that the number of traffic participants in a scene remains constant throughout the entire scenario. Therefore, no new traffic participants are added after the seed-scene, unlike some real-life recordings. Additionally, while vehicles can leave a region of interest, they continue to be simulated in the simulator. 

\subsection{Behavior Models}
\label{sec:behavior_models}
Behavioral models are utilized to assign distinct driver profiles to individual actors. Each model utilizes environmental and contextual information provided by the simulator to calculate a future trajectory.
The perception function modifies the global environmental data by narrowing the model's perspective to a specific perception area and transforms the traffic scene into a local coordinate system. 
Additionally, a downstream function converts the list of objects and HD map from the simulator into the necessary representation for the model.
This section provides a brief introduction to the models used in this work. For more detailed information, please refer to the respective publications.

\subsubsection{Path Following Model}
\label{sec:path_following_model}
The path sequence model serves as the foundation for rule-based driver profiles. This model adheres to a predetermined path, which can be specified externally, such as the actual trajectory of a vehicle in the dataset or the center of the current lane, ensuring that only the road is followed.
This model lacks the ability to avoid obstacles, and the only way to react is by adjusting speed. In such cases, the Path Following Model can be enhanced by an Intelligent Driver Model (IDM) \cite{treiber_congested_2000}.
The IDM aims to reach a target speed $v_0$ while maintaining a defined distance $s^*$ from any vehicle in front. The acceleration formula is calculated using the following equation, where $a_{\text{max}}$ defines the maximal acceleration of the vehicle, $v$ the current velocity, $b$ the comfortable braking deceleration, $s_0$ minimum distance and $s$ the headway to the vehicle driving ahead. 
The acceleration coefficient $\delta$ or different values of $s^*$ can be used to generate various driver profiles.
\begin{align}
    \label{eq:idm}
    a &= a_{\text{max}} \left(1 - \left(\frac{v}{v_0}\right)^\delta - \left(\frac{s^*}{s}\right)^2\right)\\
    s^* &= s_0 + v \cdot T + \frac{v \cdot \Delta v}{2 \sqrt{a_{\text{max}} b}} 
\end{align}

The topology information of the map is utilized to calculate the path along the road. Initially, the position of the vehicle on the road is matched to identify the first road element. Subsequently, possible routes are searched for in the road graph. If multiple routes are found, such as through a branching or intersection, they are classified in the local vehicle (in the seed-scene) coordinate system based on the angles between the start and end points of the route. This enables the same routes to be taken deterministically, depending on which parameterization is chosen to select the next route.

Vehicles should be able to pass on particularly wide roads or where a car is parked on the side of the road. As a result, only vehicles that are on or near the calculated path should be considered when driving behind them. The minimum clearance is theoretically freely configurable but is set to 5\,m in this case.

\subsubsection{Graph-based Model}
\label{sec:graph-based_model}
To enhance the realistic behavior of the vehicles, the graph-based model is based on an imitation learning approach that, unlike the IDM, uses contextual information in the form of a semantic scene graph \cite{zipfl_towards_2022} rather than just the preceding vehicle. The graph spans edges between relevant traffic participants, which contain important information such as distance along the lane, lateral offset, or type of relationship (lateral, intersecting, longitudinal). A trajectory is calculated based on the graph provided. However, it is important to note that only acceleration and deceleration are calculated, not the steering angles. A predefined path is necessary for this behavior model to follow. This path can be either predefined or calculated using the road network (see \ref{sec:path_following_model}).
For more information and analysis on this trajectory predictor, please refer to the paper \cite{zipfl_relation-based_2022}.

\subsection{Hybrid Model}
\label{sec:hybrid_model}
\begin{figure}[t]
    \centering
    \begin{minipage}[t]{0.40\linewidth}
    \includegraphics[width=\linewidth]{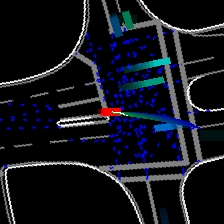}
    \end{minipage}
    \quad
    \begin{minipage}[t]{0.40\linewidth}
    \includegraphics[width=\linewidth]{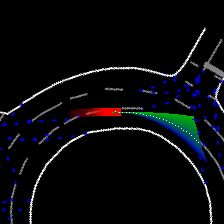}
    \end{minipage}
    \caption{Example traffic scenes and their corresponding predicted trajectory distribution are shown based on different weightings of the attention value. The weighting of the graph context information used to calculate the trajectory increases from blue to green.}
    \label{fig:example_hybrid}
\end{figure}

The architecture proposed in \cite{zipfl_utilizing_2023} employs a hybrid model that combines graph-based and image-based input data.
The main concept is to utilize contextual information of the current traffic scene, including the social interactions between traffic participants, as mapped by the semantic scene graph. Additionally, the road geometry is depicted from a top-down perspective to determine the steering angle. This approach, unlike the graph-only model (\Cref{sec:graph-based_model}), enables direct estimation of the complete trajectory within the machine-learned model.
This model has a unique feature where social and geometric information are processed in parallel and then weighted by an attention parameter before being combined. This parametrizable attention value makes it possible to make the behavior of the model more reactive (interactive) to other traffic participants, the higher the weighting of the graph component is set (see \Cref{fig:example_hybrid}).
This model allows for the user to select different behavioral profiles based on the level of the attention value.

To keep the covariate shift as small as possible, the behavior of the imitation learning models are trained on the same dataset from which the seed-scenes are derived.

\section{Experiments}
\label{sec:experiments}

The experiments aim to analyze criticalities of traffic scenes described by the road geometry and encountered traffic participants. Each traffic participant is controlled using one of the behavior models defined in \Cref{tab:behavior_models}. 
The chosen models aim to accurately reflect the behavioral spectrum of traffic scenarios and actively prompt responses from fellow traffic participants, like the model designed for emergency braking.
Machine-learned behavior models use nine additional time steps from the past for initialization, in addition to the information in the start scene. The simulation duration for a future, i.e., for a child-scenario, is 3 seconds i.e., 30 time steps. After each simulation run, the behavior models are randomly exchanged. The probability of assigning the models is uniformly distributed.

\begin{table}[tbp]
\centering
\captionsetup{justification=centering}
\caption{Models used in the Simulation}
\begin{tabular}{@{}ll@{}}
\toprule
Name                & Parametrization   \\ \midrule
Standard Driver (\ref{sec:path_following_model})        & $b$ = 3.0, $T$ = 3.1, $s_0$ = 9, $\delta$ = 4 \\
Risky Driver (\ref{sec:path_following_model})           & $b$ = 8.0, $T$ = 2.1, $s_0$ = 5, $\delta$ = 4 \\
Constant Velocity  (\ref{sec:path_following_model}) & Follows path with initial speed  \\
Emergency Brake  (\ref{sec:path_following_model})       & Decelerates abruptly at $-5.0\,\frac{m}{s^2}$       \\
Graph Predictor  (\ref{sec:graph-based_model})          & Acceleration imitation model \\
Graph Image Predictor  (\ref{sec:hybrid_model})         & Trajectory imitation model\\ \bottomrule
\label{tab:behavior_models}
\end{tabular}
\vspace{-3ex}
\end{table}

For each child-scenario, a value for each criticality metric is obtained.
Furthermore, we analyze the mean values for the extrema of the criticality metrics, as explained in \Cref{sec:metrics}. This provides further information about the progression of each scenario over time. The result is 12 values for each child-scenario, describing it based on its criticality. To aggregate all child-scenarios of a seed-scene, we accumulate the values using the respective criticality value and calculate a Kernel Density Estimation (KDE). The KDE allows for the comparison of continuous value spaces and smooths the density distribution, providing comparable results when the number of simulation runs is sampled (see \Cref{seq:sim_sampling}).
A Gaussian kernel with a bandwidth of 0.1 is used for smoothing on all metrics, respectively.

To maintain clarity in presenting the results and to conserve space, we include only selected representative examples in the figures, avoiding an overload of information in this paper.

\subsection{Sampling Size}
\label{seq:sim_sampling}

\begin{figure}
    \centering
    \begin{tikzpicture}
      \begin{axis}[
        xlabel={Distance metric value},
        ylabel={Density},
        legend entries={7776, 1000, 385, 100, 10},
        legend pos=north west,
        grid,
        no markers,
        height=5.00cm,
        width=\linewidth,
      ]
        \foreach \y in {full,1000,385,100,10} {
          \addplot table [x=x, y=\y, col sep=comma] {figures/plots/DistanceSimple_iteration_exaple.csv};
        }
      \end{axis}
    \end{tikzpicture}    
    \caption{Comparison of density functions for different sample sizes in regard to the characteristic coverage of the distance metric}
    \label{fig:plot_iterations}
    \vspace{-.5cm}
\end{figure}
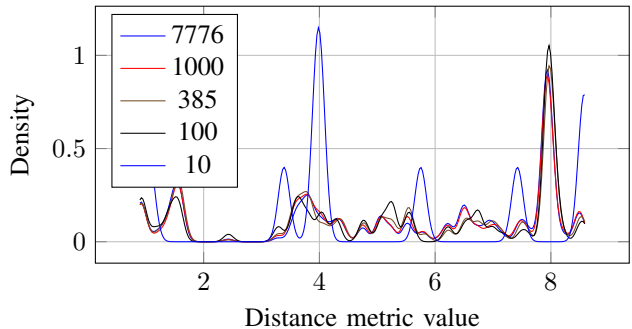

To discuss the extent to which the extrapolated future of a traffic scene affects its relevance, we first analyze the number of iterations for later simulation runs. The number of all possible futures, i.e., the different assignment of all used models to each vehicle in the scene, increases exponentially with the number of traffic participants.
Assuming an average of 11 traffic participants per scene in the dataset used and 6 different models, one would have to simulate an average of $3.6 \times 10^8$ times to calculate all possible futures in this setup. This is not feasible in terms of computing time, even for individual scenes. We will, therefore, analyze how many simulation runs are sufficient to approximate the simulation of all possible cases.
\Cref{fig:plot_iterations} illustrates the density functions of the distance metric in an exemplary scenario involving five traffic participants. To achieve this, we conducted simulations for all possible combinations of models and traffic participants, totaling 7776 runs. Subsequently, we randomly sampled scenarios of 1000, 100, and 10 types. Additionally, we specifically chose 385 runs to represent a sample size that aligns with a typical confidence level of 95\% and a margin of error of 5\%, assuming an unknown population size.

Upon examining the correspondence of the curves, it becomes apparent that a smaller sample size leads to greater deviation in the shape of the density distribution. To optimize computation efficiency, the objective is to choose the smallest feasible sample size. Consequently, all subsequent simulations were conducted with 385 runs. This allowed for the representation of significant characteristics and substantially reduced the computational time.

\subsection{Scene Criticality Analysis}
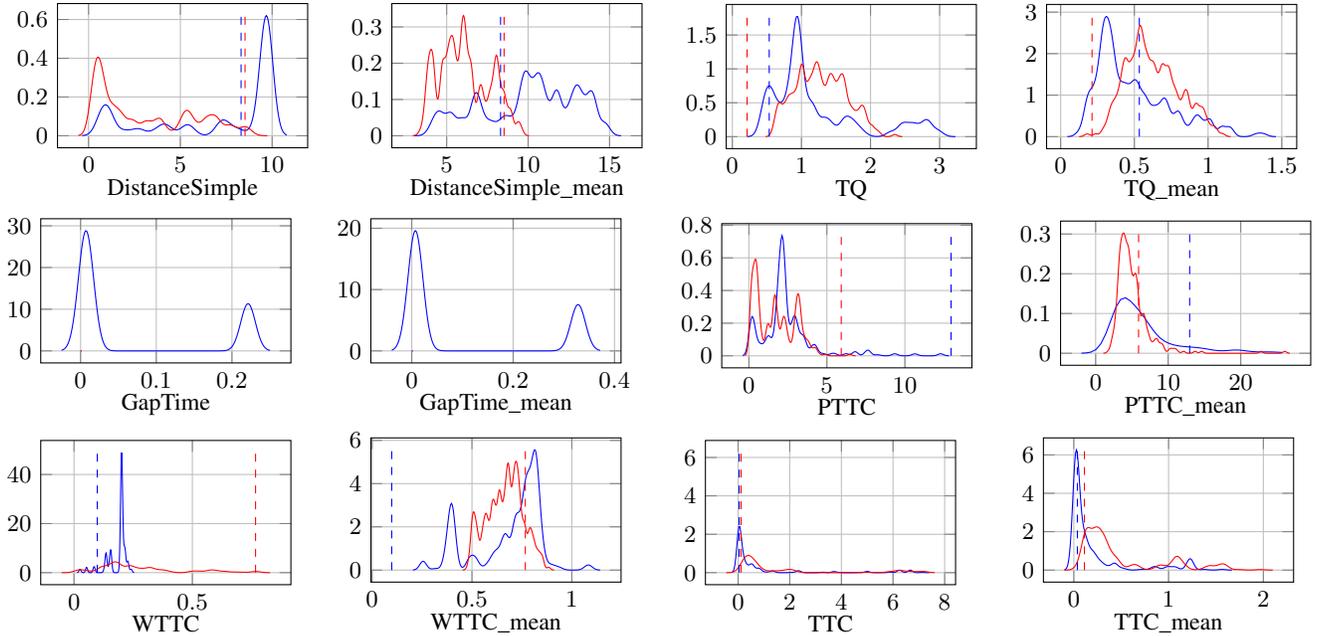
\begin{figure*}[ht]
    \
    \foreach \xcol/\ycol in {DistanceSimple/DistanceSimple, DistanceSimple_mean/DistanceSimple_mean, TQ/TQ, TQ_mean/TQ_mean, GapTime/GapTime, GapTime_mean/GapTime_mean, PTTC/PTTC, PTTC_mean/PTTC_mean,
   WTTC/WTTC, WTTC_mean/WTTC_mean,
    TTC/TTC, TTC_mean/TTC_mean} {
        \begin{minipage}[b]{0.24\textwidth}
        \centering
            \begin{tikzpicture}
                \begin{axis}[
                    baseline,
                    trim axis right,
                    xlabel shift = -1ex,
                    xlabel={\small\detokenize\expandafter{\xcol}},
                    label style={font=\small},
                    grid,
                    ticklabel style={/pgf/number format/fixed, font=\small},
                    no markers,
                    height=3.50cm,
                    width=1.15\linewidth,
                ]
                \pgfplotstableread[col sep=comma]{figures/plots/OF_col_data.csv}\OFdata
                \addplot[blue] table[x=1_\xcol_x, y=1_\ycol_y] {\OFdata};

                \pgfplotstableread[col sep=comma]{figures/plots/MT_col_data.csv}\OFdata
                \addplot[red] table[x=3_\xcol_x, y=3_\ycol_y] {\OFdata};
                \pgfplotstableread[col sep=comma]{figures/plots/OF_gt_col_data_man.csv}\OFscalarvalues
               \addplot[blue, dashed] table[x=1_\xcol_x, y=1_\ycol_y] {\OFscalarvalues};

               \pgfplotstableread[col sep=comma]{figures/plots/MT_gt_col_data.csv}\MTscalarvalues
               \addplot[red, dashed] table[x=3_\xcol_x, y=3_\ycol_y] {\MTscalarvalues};
           \end{axis}
            \end{tikzpicture}
        \end{minipage}
    }

    \caption{For all plots, the abscissa represents the criticality value and the ordinate represents the density, which is not shown in the plot for visualization purposes. The blue and red colors indicate the distribution of criticality for all child-simulations, the dashed lines marking the measured worst cases in the real situation respectively. If there is no dashed line, the metric could not be calculated for this seed-scene. Additionally, gap time could not be calculated for the red scenario.}
    \label{fig:density_plots_example}
\end{figure*}

In this subsection, we will discuss the distribution of criticality values in the simulation and in the real scenario.
\Cref{fig:density_plots_example} shows the density distributions of the analyzed metrics for two seed-scenes in blue and red.
The blue scenario was taken from a roundabout map, while the red scenario was taken from a highway entrance with a merging situation.
The ground truth, which displays the actual course of the scenario as recorded in the dataset, is marked by a dashed line.
Gap time is not applicable in the merging scenario (red) because it is a metric for intersection trajectories. In this scenario, only car following constellations exist.
In addition, some metric results have fewer than the 385 calculated values due to their inability to be computed in certain child-scenarios. 
It is noteworthy that the density distributions of the criticality distributions rarely correspond to a normal distribution. Instead, they exhibit distinct characteristics depending on the seed-scene. Additionally, it is evident that the criticality values of the ground truth are not always representative of the seed-scene. For instance, criticality metrics cannot be calculated for the ground truth scenario depicted here, but they can be computed for some child-scenarios. Additionally, the ground truth value may not necessarily fall at a specific point in the distribution, which would enable us to conclude that the simulation tends to generate more or less critical scenarios than those that would occur in reality. Instead, it seems arbitrary.

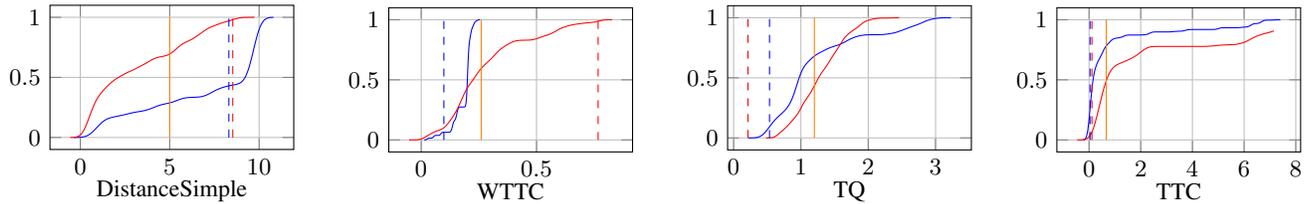
\begin{figure*}
    \centering
    \foreach \xcol/\ycol in {DistanceSimple/DistanceSimple, WTTC/WTTC, TQ/TQ, TTC/TTC} {
        \begin{minipage}{0.24\textwidth}
            \hspace*{\fill}%
            \begin{tikzpicture}
                \begin{axis}[
                    baseline,
                    trim axis left,
                    xlabel shift = -1ex,
                    xlabel={\small\detokenize\expandafter{\xcol}},
                    label style={font=\small},
                    legend pos=north west,
                    grid,
                    ticklabel style={/pgf/number format/fixed, font=\small},
                    no markers,
                    height=3.50cm,
                    width=1.13\linewidth,
                ]
                \pgfplotstableread[col sep=comma]{figures/plots/OF_cumulative_col_data.csv}\OFdata
                \addplot[blue] table[x=1_\xcol_x, y=1_\ycol_y] {\OFdata};

                \pgfplotstableread[col sep=comma]{figures/plots/MT_cumulative_col_data.csv}\OFdata
                \addplot[red] table[x=3_\xcol_x, y=3_\ycol_y] {\OFdata};
                \pgfplotstableread[col sep=comma]{figures/plots/OF_cumulative_gt_col_data.csv}\OFscalarvalues
               \addplot[blue, dashed] table[x=1_\xcol_x, y=1_\ycol_y] {\OFscalarvalues};

               \pgfplotstableread[col sep=comma]{figures/plots/MT_cumulative_gt_col_data.csv}\MTscalarvalues
               \addplot[red, dashed] table[x=3_\xcol_x, y=3_\ycol_y] {\MTscalarvalues};

               \pgfplotstableread[col sep=comma]{figures/plots/thresholds.csv}\thresholds
               \addplot[orange] table[x=1_\xcol_x, y=1_\ycol_y] {\thresholds};
           \end{axis}
        \end{tikzpicture}
        \end{minipage}
    }
    \caption{For all plots, the abscissa represents the criticality value and the ordinate represents the density, which is not shown in the plot for visualization purposes. The plot shows cumulative density plots for the criticality metrics distance, WTTC, traffic quality, and TTC. The blue and red curves describe the same simulations as in \Cref{fig:density_plots_example}. The orange line indicates the suggested criticality threshold from literature.}
    \label{fig:density_plots_example_cumulative}
    \vspace{-0.3cm}
\end{figure*}

\Cref{fig:density_plots_example_cumulative} describes the cumulative density for selected metrics of both seed-scenes from \Cref{fig:density_plots_example}. The orange line indicates the suggested criticality threshold for all metrics in the literature. In case of distance and WTTC all simulations on the left side of the line are critical simulations. For the traffic quality and the inverted TTC all simulations on the right are critical. We used the following thresholds: distance 5\,m, WTTC 0.26\,s (physical limit to brake), traffic quality 1.2 (derived from evaluations in \cite{schutt_inverse_2023}), TTC 1.5\,s.
In terms of distance, the red scenario is the more critical seed-scene, as the curve is faster growing than the blue one. However, this can be explained by shortcomings in the distance measure. Drivers often travel side by side in different lanes during merging. This results in small distances and therefore high criticality. Additionally, the situation would not be considered critical by human drivers.
The TTC metric for the red curve also shows a more critical scene than the blue curve. Again, this is because TTC can only be calculated for car-following constellations, and the red seed-scene only consists of car-following combinations.
For WTTC, the blue curve shows a slightly more critical seed-scene in \Cref{fig:density_plots_example} and \Cref{fig:density_plots_example_cumulative}. The blue curve reaches 1 before the threshold, indicating that all simulated child-scenarios have a lower WTTC value than the considered threshold. Unlike classical TTC, WTTC can be computed in any constellation and can therefore find critical situations that TTC would miss.
In addition, the WTTC has a tendency to classify more situations as critical as there actually are \cite{wachenfeld_worst-time--collision_2016}. The aim in designing this metric was to label non-critical scenes as critical, preferably to overlook critical scenes.
When evaluating traffic quality, the red seed-scene is slightly more critical but grows faster in the upper quarter.
Generally, seed-scenes with critical child-scenarios on the left side of the threshold are more significant as the curve becomes steeper. Conversely, seed-scenes with critical child-scenarios on the right side are more significant when the curve starts to grow on the right side.

\section{Conclusion and Outlook}
\label{sec:conclusion}
In the scope of this work, a framework was developed that allows the presentation of scenes that can be used as initialization for a test scenario for a HADF. 
Each action and behavior of a traffic participant within this scene can change the outcome of the scenario. Therefore, different futures (child-scenarios) were simulated to get a broader understanding of the scene. 
An explicit investigation with a driving function to be tested was intentionally not carried out, as we want to make a holistic and objective statement about the entire scene and not for a specific ego-vehicle.

The experiments utilized behavior models chosen to reproduce as many interactions as possible while remaining true to reality. However, the quality of the selected behavior models directly impacts the results. In some experiments, the traffic participant controlled by the hybrid imitation behavior model left the lane and did not find its way back to the road. This phenomenon is an example of the fact that the simulations performed cannot provide a definitive assessment of a traffic scene. However, we believe that by depicting various eventualities, even if only a subset of all possible futures, we can significantly improve our understanding of the situation.
Furthermore, we evaluated the outcome of all child scenarios from one seed-scene regarding their criticality with different criticality metrics in our experiments.
Additionally, we showed that different seed-scenes can lead to different criticality characteristics (e.g., merging has a higher criticality regarding distance).
Future work should build on the feature space resulting from the simulation to group and classify scenes. The density distributions of each traffic scene provide a distinct fingerprint for each scene, which could be used to identify different features relevant for testing.

\bibliographystyle{IEEEtran}
\bibliography{references_cleaned}
\end{document}